\documentclass[letterpaper]{article} % DO NOT CHANGE THIS
\usepackage{aaai2026}  % DO NOT CHANGE THIS #my change this arxiv
\usepackage{times}  % DO NOT CHANGE THIS
\usepackage{helvet}  % DO NOT CHANGE THIS
\usepackage{courier}  % DO NOT CHANGE THIS
\usepackage[hyphens]{url}  % DO NOT CHANGE THIS
\usepackage{graphicx} % DO NOT CHANGE THIS
\usepackage{natbib}  % DO NOT CHANGE THIS AND DO NOT ADD ANY OPTIONS TO IT
\usepackage{caption} % DO NOT CHANGE THIS AND DO NOT ADD ANY OPTIONS TO IT
\usepackage{algorithm}
\usepackage{algorithmic}

\usepackage{subcaption}
\usepackage{makecell}
\usepackage{multirow}
\usepackage{booktabs}
\usepackage{amsmath}
\usepackage[most]{tcolorbox}
\definecolor{egggreen}{HTML}{d7e6e8}
\definecolor{lightskyblue3}{HTML}{dbdbeb}
\definecolor{lightyellowgreen3}{HTML}{d6ecf0}
\definecolor{constraint}{HTML}{eee2f0}
\definecolor{LemonChiffon2}{HTML}{EEE9BF}
\definecolor{Ivory2}{HTML}{EEEEE0}
\definecolor{LavenderBlush2}{HTML}{EEE0E5}
\definecolor{RED}{HTML}{FCBBC3}

\newtcolorbox{mybox}[1][]{
	width=\columnwidth,
	colback = gray!6, 
	colframe = black, 
	boxrule = 0.8pt,
	boxsep=0pt,left=10pt,right=10pt,top=8pt,bottom=8pt,
	fontupper=\linespread{1.2}\selectfont,
	title=#1}

\usepackage{newfloat}
\usepackage{listings}
\DeclareCaptionStyle{ruled}{labelfont=normalfont,labelsep=colon,strut=off} % DO NOT CHANGE THIS
\floatstyle{ruled}
\newfloat{listing}{tb}{lst}{}
\floatname{listing}{Listing}
\nocopyright %-- Your paper will not be published if you use this command
\title{In-Context Examples Matter: Improving Emotion Recognition in Conversation with Instruction Tuning}

\author{
    Hui Ma\textsuperscript{\rm 1}, Bo Zhang\textsuperscript{\rm 2}, Jinpeng Hu\textsuperscript{\rm 1}, Zenglin Shi\textsuperscript{\rm 1}\thanks{Corresponding Author: zenglin.shi@hfut.edu.cn}%,
}
\affiliations{
    \textsuperscript{\rm 1}Hefei University of Technology\\
    \textsuperscript{\rm 2}Zhengzhou University\\
}

\usepackage{bibentry}
\begin{document}

\maketitle

\begin{abstract}
% Recently, large language models (LLMs) have shown significant promise for the ERC task. 
% Large language models (LLMs) have shown impressive results/performance on ERc task. gradually become a key focus in ERC. 
Emotion recognition in conversation (ERC) aims to identify the emotion of each utterance in a conversation, playing a vital role in empathetic artificial intelligence. With the growing of large language models (LLMs), instruction tuning has emerged as a critical paradigm for ERC. Existing studies mainly focus on multi-stage instruction tuning, which first endows LLMs with speaker characteristics, and then conducts context-aware instruction tuning to comprehend emotional states. However, these methods inherently constrains the capacity to jointly capture the dynamic interaction between speaker characteristics and conversational context, resulting in weak alignment among speaker identity, contextual cues, and emotion states within a unified framework. In this paper, we propose \textbf{InitERC}, a simple yet effective one-stage \textbf{In}-context \textbf{i}nstruction \textbf{t}uning framework for \textbf{ERC}. InitERC adapts LLMs to learn speaker-context-emotion alignment from context examples via in-context instruction tuning. Specifically, InitERC comprises four components, i.e., demonstration pool construction, in-context example selection, prompt template design, and in-context instruction tuning. To explore the impact of in-context examples, we conduct a comprehensive study on three key factors: retrieval strategy, example ordering, and the number of examples. Extensive experiments on three widely used datasets demonstrate that our proposed InitERC achieves substantial improvements over the state-of-the-art baselines.
\end{abstract}


\section{Introduction}
% wide applications  promising potential Distinct from
% Unlike jointly aiming opinion mining, 
Emotion recognition in conversation (ERC) is a fundamental task in natural language processing (NLP) that aims to identify the emotion of each utterance in a conversation. The task has been popularly explored due to its potential applications in recommendation \cite{song2024cagk}, social media analysis \cite{brambilla2022graph}, and empathetic dialogue systems \cite{zhou2020design}. Unlike sentence-level emotion recognition \cite{kim-2014-convolutional,deng2023llms,zhang2024sentiment}, ERC necessitates modeling speaker characteristics and contextual dependencies, as both are essential for understanding emotional states within the hierarchical and interactive structure of conversations.

% conversational context

% two complementary sources of information that are essential for capturing the nuanced emotional dynamics inherent in dialogue structure.

% and contextual dependencies, as these factors are essential for understanding emotional expressions embedded within the hierarchical and interactive structure of conversations.

% ERC 需要建模说话者和上下文信息，由于对话结构xxx

% ERC necessitates

% 传统的ERC研究主要分为两类. The former the latter
% mainstream 
% 按照下面的思路将下面这段话补充完整，随后可以进行改写，使得更加科学 严谨 合理 流畅和学术化
Previous works on ERC can be generally divided into two categories \cite{li2023skier}: sequence-based methods and graph-based methods. Sequence-based methods, such as HiTrans \cite{li2020hitrans}, MVN \cite{ma2022multi}, BERT-ERC \cite{qin2023bert}, and SDT \cite{ma2024sdt}, typically model conversational utterances as temporal sequences and leverage recurrent neural networks (RNNs) or transformer-based architectures to capture contextual dependencies. Graph-based methods, including SKAIG \cite{li2021past}, DAG-ERC \cite{shen2021directed}, DualGATs \cite{zhang2023dualgats}, and ESIHGNN \cite{zha2024esihgnn}, construct conversational graphs and employ graph neural networks (GNNs) to capture dependencies and interactions within the dialogue structure. Despite significantly enhancing emotion recognition by leveraging contextual and speaker-specific information, these methods heavily rely on manually crafted graph structures and intricate model designs.

More recently, the advent of large language models (LLMs) has marked a paradigm shift in NLP research. Pre-trained on massive and diverse corpora, LLMs have demonstrated impressive performance on a wide range of NLP tasks \cite{lin2024data,li2025exploring,yuan2025query}, including ERC. A mainstream strategy for effectively adapting LLMs to ERC is instruction tuning, which integrates task-specific rules during fine-tuning to achieve significant performance improvements. BiosERC \cite{xue2024bioserc} extracts speaker biographical information using LLMs as supplementary knowledge to fine-tune LLMs. To enhance emotion understanding, InstructERC \cite{lei2023instructerc} and LaERC-s \cite{fu2025laerc} adopt a two-stage instruction tuning framework. In the initial stage, speaker identification or characteristic instruction tuning is performed to endow LLMs with speaker characteristics. In the second stage, context-aware instruction tuning is conducted to comprehend emotional states within conversational contexts. However, these sequential methods inherently constrain the capacity to jointly capture the dynamic interplay between speaker characteristics and conversational context, leading to weak alignment among speaker identity, contextual cues, and emotion states.
%LLMs such as GPT, LLaMA, and Mistral have demonstrated impressive performance on a wide range of NLP tasks, including ERC.
% within a unified framework. 

In this paper, we propose InitERC, a simple yet effective one-stage in-context instruction tuning framework for ERC. InitERC enables LLMs to learn the alignment among speaker identity, conversational context, and emotional state from context examples via instruction tuning. Specifically, InitERC consists of four components: demonstration pool construction creates a candidate set of context- and speaker-aware examples; in-context example selection retrieves the most relevant demonstration examples for each target input; prompt template design integrates the task description, target input, and selected in‐context examples into a unified prompt format; and in-context instruction tuning adapts the LLM to capture speaker-context-emotion alignment through a single-stage instruction tuning with the constructed prompts. Furthermore, we conduct a comprehensive study on retrieval strategy, example ordering, and the number of examples to explore the effects of in-context examples, and we hope our findings will offer useful insights for future research on in-context instruction tuning for ERC.
% In summary, our main contributions are as follows:
% The contri- butions of our work are summarized as follows
Our main contributions are summarized as follows:
\begin{itemize}
    \item We propose InitERC, a one-stage in-context instruction tuning framework for ERC, which enables LLMs to learn the alignment among speaker identity, conversational context, and emotional state from context examples via instruction tuning.
    \item We conduct a comprehensive study on retrieval strategy, example ordering, and the number of examples to investigate how these factors influence the effectiveness of in-context examples. % for our InitERC
    \item Extensive experiments on IEMOCAP, MELD, and EmoryNLP datasets demonstrate the effectiveness of our proposed InitERC framework, achieving state-of-the-art performance.
    % Experimental results
    % demonstrate
    % We conduct extensive experiments on three benchamark datasets, and the results show the effectiveness of our framework, achieving state-of-the-art performances.
    % Experimental results on three benchamark datasets show the effectiveness of our framework, achieving state-of-the-art performances.
    % Experimental results on three benchamark datasets show that our framework exhibits xx, achieving state-of-the-art performances.
    % superior performance.
    % achieves state-of-the-art performance on the ERC task
\end{itemize}

\section{Related Work}
\subsection{Emotion Recognition in Conversation}
% Conventional ERC studies are mainly divided into two categories: sequence-based mehtods and graph-based methods. 
% \cite{li2020hitrans,ma2022multi,shen2021dialogxl,shen2021dialogxl,qin2023bert}
% \cite{ghosal-etal-2019-dialoguegcn,zhang2023dualgats,zha2024esihgnn} 
Conventional ERC studies are mainly divided into sequence-based methods and graph-based methods \cite{li2023skier}. Sequence-based methods use RNNs or transformer-based architectures to capture contextual dependencies. HiTrans \cite{li2020hitrans} uses two hierarchical transformers to capture hierarchical context information. DialogXL \cite{shen2021dialogxl} extends XLNet \cite{yang2019xlnet} with enhanced memory and dialog-aware self-attention to model long-range multi-party dialogues. While graph-based methods employ GNNs to capture dependencies within the dialogue structure. DAG-ERC \cite{shen2021directed} introduces a directed acyclic graph neural network to capture the temporal and speaker-aware contextual information. DualGATs \cite{zhang2023dualgats} leverages dual graph attention networks to jointly model discourse structure and speaker-aware context. 

With the development of LLMs, several works have explored their potential for ERC. BiosERC \cite{xue2024bioserc} fine-tunes LLMs using extracted speaker biographical information as supplementary knowledge. InstructERC \cite{lei2023instructerc} and LaERC-S \cite{fu2025laerc} adopt a two-stage instruction tuning framework, where first stage equips LLMs with speaker-specific characteristics, followed by context-aware instruction tuning to interpret emotional states. However, such sequential frameworks limit the ability to jointly model the dynamic interaction between speaker characteristics and conversational context, leading to suboptimal alignment among speaker identity, contextual cues, and emotional states. In this paper, we propose InitERC, a one-stage instruction tuning framework, to learn speaker-context-emotion alignment from context examples via instruction tuning. %in-context 

\subsection{Instruction Tuning} % one
Instruction tuning \cite{shengyu2023instruction}, also known as supervised fine-tuning, is an effective technique for enabling LLMs to follow natural language instructions and perform a wide range of tasks. It involves further training LLMs to generate certain outputs given instruction prompts. Recent studies, including InstructionGPT \cite{ouyang2022training}, OPT-IML \cite{iyer2022opt}, and FLAN \cite{chung2024scaling}, have systematically explored instruction tuning methods to enhance the capabilities and controllability of LLMs. In addition to serving as generalists, instruction tuning can also function as specialists \cite{lee2024instruction}, excelling in specific tasks rather than achieving proficiency across a broad range of tasks. For example, Chen et al. \shortcite{cheng2024emotion} employed a modified LLaMA model with instruction tuning to boost not only the accuracy of emotion recognition but also the depth of emotional reasoning. Jian et al. \shortcite{jian2025simrp} addressed the aspect sentiment quad prediction task by designing prompts with syntactic and semantically similar demonstrations. This word constructs effective instruction prompt templates and focuses on instruction tuning for ERC task.

\subsection{In-Context Learning} 
In‐context learning (ICL) \cite{dong2024survey} has recently gained increasing attention as a novel training-free paradigm for enhancing the performance of LLMs. The approach enables LLMs to perform diverse tasks using a few input-output examples as demonstrations, without requiring retraining or fine-tuning. For instance, GPT-3 \cite{brown2020language} demonstrates competitive performance on unseen tasks when provided with few-shot examples as input prefixes, achieving results comparable to fine-tuned models. To further improve the effectiveness of ICL, some studies have investigated integrating it with fine-tuning. Chen et al. \shortcite{chen2022meta} proposed in-context tuning, which meta-trains language models with a few examples. DiSTRICT \cite{venkateswaran2023district} adopts in-context tuning by fine-tuning language models with in-context examples for dialogue state tracking. Chen et al. \shortcite{chen2025reactgpt} enhanced the performance of LLMs on chemical reaction captioning task via in-context tuning. Similarly, Li et al. \shortcite{li2025large} guided LLMs to derive knowledge from informative context examples in the molecule-caption translation task. In this work, we introduce in-context instruction tuning to improve the performance of LLMs on ERC task.
% some studies explore combining it with fine-tuning.
% GPT-3 \cite{brown2020language} could achieve performance on unseen tasks competitive with fine-tuned models when provided with few-shot examples as a prefix. 
 % leveraged informative context examples to guide LLMs in the molecule-caption translation task.

% A survey on in-context learning

%  GPT 3 Language models are few-shot learners

% These strong abilities have been widely verified as emerging abilities for LLMs.

% , highlighting emergent ICL capabilities at scale.
% A notable demonstration of this capability is GPT-3, which achieves results on unseen tasks comparable to fine-tuned models by using a small number of examples as a prefix (Brown et al., 2020).
% enabling LLMs to perform various tasks using a few input-output examples as demonstrations. 

% In‐context learning (ICL) has recently gained attention as a novel learning-free paradigm for enabling LLMs to perform various tasks using a few input-output examples as demonstrations. 

% Many studies have shown that LLMs
% can perform a series of complex tasks through
% ICL,

% This paradigm enables LLMs to perform diverse tasks by simply conditioning on examples, without additional training.

% However, the organization of these sample-label
% pairs is critical, as it can significantly affect ICL
% performance1 (Liu et al., 2022).

% However, the organization of these sample-label
% pairs is critical, as it can significantly affect ICL
% performance1 (Liu et al., 2022).

\section{Methodology}
\subsection{Task Definition}% /problem formulation
% each utterance $u_t$ is uttered by one of the
% A conversation $\mathcal{C} = \left\{ \left(s_{u_1}, u_1 \right), \left(s_{u_2}, u_2 \right), \cdots, \left(s_{u_N}, u_N \right)\right\}$ consists of $N$ consecutive utterances and each utterance $u_t$ is uttered by $s_{u_t}$, where $s$ maps the utterance into its corresponding speaker. The ERC task aims to predict the emotion label $y_t$ for each utterance $u_t$ based on its historical context $c_t = \left\{ \left(s_{u_1}, u_1 \right),\left(s_{u_1}, u_2 \right),\cdots,\left(s_{u_{t-1}}, u_{t-1} \right) \right\}$ and its speaker $s_{u_t}$. The emotion label belongs to the pre-defined emotion category set $\left\{ {{e_1},{e_2},\cdots,{e_K}} \right\}$ and $K$ is the number of emotion categories.

\begin{figure*}[t]
\centering %0.95
\includegraphics[width=1.0\textwidth]{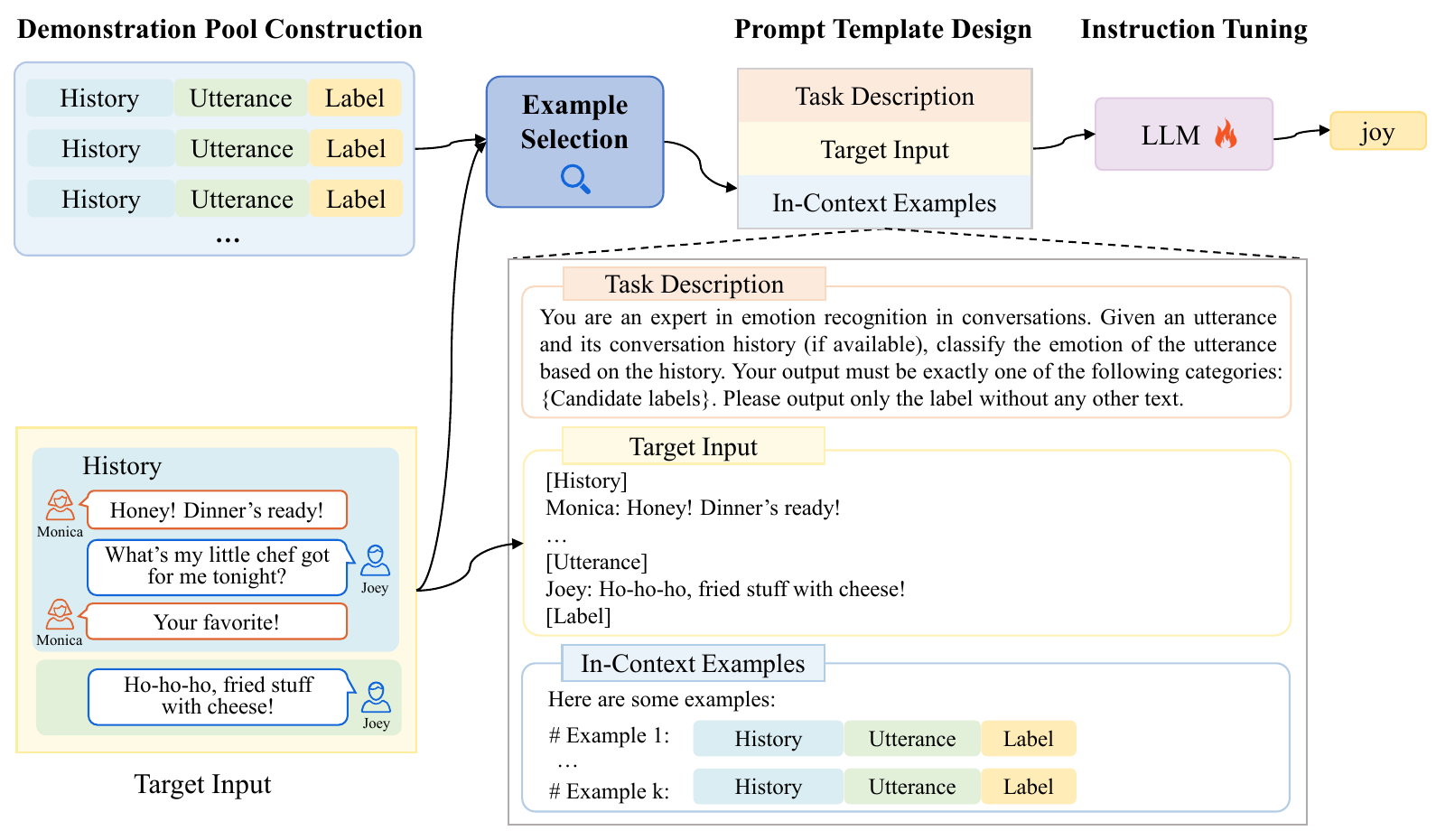} % Reduce the figure size so that it is slightly narrower than the column.
\caption{Framework of the proposed InitERC.}
\label{initerc}
\end{figure*}

A conversation consists of $N$ consecutive utterances $\left\{ {{u_1},{u_2},\cdots,{u_N}} \right\}$ uttered by a sequence of speakers $\left\{ s_{u_1},s_{u_2}, \cdots ,s_{u_N} \right\}$, where $s$ maps each utterance into its corresponding speaker. The ERC task aims to predict the emotion label $y_t$ of each utterance $u_t$ based on its historical context $c_t = \left\{ \left(s_{u_1}, u_1 \right),\left(s_{u_1}, u_2 \right),\cdots,\left(s_{u_{t-1}}, u_{t-1} \right) \right\}$ and its speaker $s_{u_t}$. The emotion label belongs to the pre-defined emotion category set $\left\{ {{e_1},{e_2},\cdots,{e_M}} \right\}$ and $M$ is the number of emotion categories.
% 
% A conversation consists of $N$ consecutive utterances $\left\{ {{u_1},{u_2},\cdots,{u_N}} \right\}$ and $M$ speakers $\left\{ {{s_1},{s_2}, \cdots ,{s_M}} \right\}$. Each utterance $u_t$ is spoken by the speaker ${s_{\phi \left( {u_t} \right)}}$, where $\phi $ maps each utterance into its corresponding speaker. The ERC task aims to predict the emotion label $y_t$ for each utterance $u_t$ based on its historical context $c_t = \left\{ \left(u_1 \right),{u_2},\cdots,u_{t-1}\right\}$ and its speaker ${s_{\phi \left( {u_t} \right)}}$. The emotion label belongs to the pre-defined emotion category set $\left\{ {{e_1},{e_2},\cdots,{e_K}} \right\}$ and $K$ is the number of emotion categories.
% The ERC task aims to predict the emotion label $y_t$ for each utterance $u_t$ based on its historical context $c_t = \left\{ {{u_1},{u_2},\cdots,{u_{t-1}}} \right\}$ and speaker information. The emotion label belongs to the pre-defined emotion category set $\left\{ {{e_1},{e_2},\cdots,{e_K}} \right\}$ and $K$ is the number of emotion categories.
% the index of the utterance into that of the corresponding speaker
% is the mapping between an utterance and its corresponding speaker's index. 
\subsection{InitERC Overview} % consists of
The architecture of our proposed InitERC framework is shown in Figure~\ref{initerc}. InitERC integrates four components: demonstration pool construction, which builds a context‑ and speaker‑aware candidate example set, in-context example selection, which retrieves the most relevant demonstration examples; prompt template design, which integrates the task description, target input, and in‐context examples into a unified prompt format; and in-context instruction tuning, which enables the LLM to learn speaker-context-emotion alignment in a one-stage training process using the constructed prompts.
% demonstration pool construction for building a context‑ and speaker‑aware candidate set, in-context example selection for retrieving the most relevant demonstration examples, prompt template design for integrating clear task description, target input, and in‐context examples into a unified prompt format; and in-context instruction tuning for enabling LLMs to learn speaker-context-emotion alignment in a single-stage training process using the constructed prompts.
%总结一句！！！ InitERC uses xx能够xx.
% , including demonstration pool construction, 
% \begin{figure}[t]
% \centering
% \includegraphics[width=0.9\columnwidth]{figure1} % Reduce the figure size so that it is slightly narrower than the column. Don't use precise values for figure width.This setup will avoid overfull boxes.
% \caption{Using the trim and clip commands produces fragile layers that can result in disasters (like this one from an actual paper) when the color space is corrected or the PDF combined with others for the final proceedings. Crop your figures properly in a graphics program -- not in LaTeX.}
% \label{fig1}
% \end{figure}

\subsection{Demonstration Pool Construction}
Given an ERC dataset, we construct a demonstration pool $\mathcal{P}$ using its training set. In ERC, each utterance's emotional state is not only influenced by its content but also highly dependent on the preceding conversational context and the identity of the speaker. The historical context provides essential background information, such as conversational flow and emotional dependencies, while speaker identity helps to understand personalized emotional expressions and discourse patterns. Therefore, it is critical that each demonstration example includes the conversational history, current utterance and its speaker identity to ensure context- and speaker-aware emotion recognition.\par
Formally, the demonstration pool is defined as $\mathcal{P}=\left\{ \left( c_i, x_i, y_i\right) \right\}_{i=1}^K$, where $c_i$ is the historical context of utterance $u_i$, $x_i = \left(s_{u_i}, u_i\right)$ contains utterance $u_i$ and its speaker $s_{u_i}$, $y_i$ is the corresponding emotion label of utterance $u_i$, and K is the total number of examples. This carefully constructed demonstration pool serves as the candidate set for subsequent in‑context example selection, ensuring that each example provides rich contextual and speaker-specific cues to guide instruction tuning.
% each demonstration example is a triple consisting of the historical context, the speaker-aware utterance, and its corresponding emotion label. Specifically, 
% Given an ERC dataset, we construct a demonstration pool $\mathcal{P}$ using its training set. Due to both historical context and speaker information are important characters in shaping the emotional state of an utterance in conversation, each demonstration example is designed to incorporate both the conversational history and the speaker identity. Formally, the demonstration pool $\mathcal{P}=\left\{ \left( c_i, x_i, y_i\right) \right\}_{i=1}^K$ consists of (history, utterance, label) triples, where $c_i$ is the historical context of utterance $u_i$, $x_i = \left(s_{u_i}, u_i\right)$ contains utterance $u_i$ and its speaker, $y_i$ is the corresponding emotion label of utterance $u_i$, and K is the total number of utterances. This carefully constructed demonstration pool serves as the candidate set for subsequent in‑context example selection.%$\mathcal{P}$ 

\subsection{In-Context Example Selection}%/Retrieval 
For a given target input comprising the query utterance $u_t$ and its speaker $s_{u_t}$, denotes as $x_t$, along with its historical context $c_t$, we employ the off-the-shelf retriever to select the $k$ most relevant examples $\mathcal{P}^e =\left\{ \left( c_i, x_i, y_i\right) \right\}_{i=1}^k$ from the demonstration pool $\mathcal{P}$:
\begin{equation}
    \mathcal{P}^e = Retriever\left( c_t \oplus x_t, \mathcal{P} \right),
\end{equation}
% using off-the-shelf retriever. Specifically, 
where $\oplus$ denotes the concatenation operation.\par
In this work, we adopt Contriever-MS MARCO \cite{izacardunsupervised} as our retriever that is a dual transformer encoder architecture and calculates the similarity between target input and candidates via dot product. Contriever-MS MARCO is pre-trained using MoCo contrastive loss \cite{he2020momentum} on unsupervised data and further fine-tuned on the MS MARCO dataset. \par
During retrieval, we exclude from the demonstration pool any examples originating from the same dialogue as the query utterance to prevent information leakage. Moreover, emotion labels are not used during retrieval and are incorporated afterward when constructing the final in-context examples, thereby ensuring that retrieval remains label-agnostic and mitigating potential bias toward specific emotional categories.

% Moreover, emotion labels are not used during retrieval and are incorporated afterward when constructing the final in-context examples. This design ensures that retrieval remains label-agnostic and mitigates potential bias toward specific emotional categories.
% only appended after retrieval to construct the final in-context examples. 
% This design ensures that retrieval remains label-agnostic and avoids biasing selection toward particular emotional categories.
% we do not utilize the emotion labels during retrieval. The labels are only appended after retrieval to construct the final in-context examples. This design choice ensures that retrieval remains label-agnostic xx
% we do not utilize the emotion labels during retrieval. The labels are only appended after retrieval to construct the final in-context examples. This design choice ensures that retrieval remains label-agnostic xx
% To ensure fair and unbiased selection, we exclude from the demonstration pool any examples originating from the same dialogue as the query utterance.
% and aligns with realistic test-time constraints where emotion labels are not available during input construction.
% This prevents potential information leakage and avoids overfitting to specific conversation threads, ensuring that the retrieved examples are both contextually relevant and independent.
% During retrieval, emotion labels of demonstration examples are not used due to xx. 
% the emotion labels in the demonstration pool are not used due to xx. 
%需要解释原因，例如为了和target input保持一致

\subsection{Prompt Template Design}%Instruction/
% directly determines how the task is presented to LLMs and 
Prompt template serves as a critical bridge between LLMs and the ERC task. In our InitERC framework, the instruction prompt not only defines the task in a clear and explicit manner but also effectively guides the LLM to align speaker identity, conversational context, and emotional state within a unified framework. The instruction prompt template comprises three parts, i.e., task description $\mathcal{T}$, target input $\mathcal{I}$, and in-context examples $\mathcal{E}$.
\begin{itemize}
%先描述Task Description的作用
\item \textbf{Task Description} $\mathcal{T}$ provides explicit instructions that define the objective of the task and the expected output format. The task description $\mathcal{T}$ is designed as:
\begin{tcolorbox}[
    enhanced,
    breakable,
    colback=white,
    colframe=black,
    boxrule=0.2mm, 
    left=1mm,           % 左缩进2mm（默认4mm）
    right=1mm,          % 右缩进2mm（默认4mm）
    % title=Task Description $\mathcal{T}$,
    coltitle=white,
    fonttitle=\bfseries,
    listing only,
    listing options={
      basicstyle=\ttfamily\footnotesize,
      breaklines=true,
      showstringspaces=false
    }
]
You are an expert in emotion recognition in conversations. Given an utterance and its conversation history (if available), classify the emotion of the utterance based on the history. Your output must be exactly one of the following categories: $\left\{ \textnormal{Candidate labels} \right\}$. Please output only the label without any other text.
\end{tcolorbox} %``gpt-3.5-turbo''
\item \textbf{Target Input} $\mathcal{I}$ embeds the recognized speaker-aware utterance $x_t = \left(s_{u_t}, u_t\right)$ into the prompt together with its historical context $c_t$, so that LLMs can leverage both ``who said what'' and ``what was said before''. The target input is defined as:
% embeds the specific recognized utterance $u_t$ into the prompt together with its speaker $s_{u_t}$ and historical context $c_t$, so that LLMs can leverage both ``who said what'' and ``what was said before''. The target input is defined as:
% contains
% builds on the recognized utterance $u_t$ containing not only the utterance $u_t$ but also its speaker $s_{u_t}$ and historical context $c_t$, hence LLMs could xx. 
\begin{equation*}
    \mathcal{I}(u_t)=``[\textnormal{History}]\oplus c_t\oplus [\textnormal{Utterance}]\oplus x_t \oplus [\textnormal{Label}]".
\end{equation*}
% where $x_t = \left(s_{u_t}, u_t\right)$.
\item \textbf{In-context Examples} $\mathcal{E}$ are selected from in-context example selection that contain $k$ most relevant examples $\left\{ \left( c_i, x_i, y_i\right) \right\}_{i=1}^k$, and we adopt a similarity‑descending order, i.e., similar-first, for example ordering. The in-context examples is constructed as:%The examples prime LLMs to align speaker cues, contextual patterns, and labels.
\begin{align*}
    \mathcal{E}=&``\textnormal{Here are some examples:}\\
    &[\textnormal{History}]\oplus c_1\oplus [\textnormal{Utterance}]\oplus x_1 \oplus [\textnormal{Label}]\oplus y_1\\
    &[\textnormal{History}]\oplus c_2\oplus [\textnormal{Utterance}]\oplus x_2\oplus [\textnormal{Label}]\oplus y_2\\
    &\cdots\\
    &[\textnormal{History}]\oplus c_k\oplus [\textnormal{Utterance}]\oplus x_k \oplus [\textnormal{Label}]\oplus y_k".
\end{align*}
\end{itemize}

\subsection{In-Context Instruction Tuning}
%解决xx问题，用in-context instruction tuning
% to fine-tune LLM to learn from demonstration xx examples and current targe input.
% ,  which guides LLMs learn the speaker-context-emotion alignment from context examples
To overcome the limitations of multi-stage tuning and enable unified learning of speaker traits, contextual cues, and emotion states, we introduce in‑context instruction tuning. Formally, given the training dataset $\mathcal{D}$ and prompt template $\left\{ \mathcal{T, I, E}\right\}$, let the recognized utterance be $u_t$ and its emotion label be $y_t$, where $(u_t, y_t) \in \mathcal{D}$ denotes the utterance-label pair. The learning objective is to minimize the negative log-likelihood loss:
\begin{equation}
    \mathcal{L}(\theta) = -\sum_{\left (u_t,y_t \right) \in \mathcal{D}} \log p_{\theta}\left (y_t \mid \mathcal{T}, \mathcal{I}(u_t), \mathcal{E} \right),
\end{equation}
where $\theta$ denotes the trainable parameters in LLM.\par
InitERC enables LLMs learn the speaker-context-emotion alignment from context examples in a single learning stage. During inference, the demonstration examples for the recognized utterance are also selected from the constructed demonstration pool.
% InitERC furnishes the LLM with both clear instructions and targeted examples, enabling it to learn speaker–context–emotion alignment in one unified in‑context tuning stage.
% 最后总结InitERC作用！通过学习xx，有啥作用。
% By unifying instruction and example learning in a single in‑context tuning step, 
% LLMs learn the speaker-context-emotion alignment from context examples
% This structured prompt design facilitates coherent and effective in-context instruction tuning by providing a clear task definition, rich contextualized input, and high-quality demonstrations. It enables the LLM to better model the alignment among speaker identity, conversational context, and emotional state, thereby improving its ERC performance.
\section{Experimental Settings}
% \subsection{Datasets and Evaluation Metrics}
\subsection{Datasets and Evaluation Metrics}
We evaluate our framework on three benchmark datasets. Statistics of these datasets are listed in Table~\ref{tab:statics}.

% 表格局部使用 9pt Roman 字体，无额外行距
{
% \fontsize{10}{12}\selectfont  % 字体大小和行距均为 9pt（紧凑排版）
% \rmfamily  
\setlength{\tabcolsep}{1mm}  % 设置列之间的水平间距为 1mm（紧凑）
\begin{table}[t]
% \scriptsize
% % \footnotesize  
% % \small
\centering
\renewcommand\arraystretch{1.25}
\begin{tabular}{cccccccccc}
\toprule
\multirow{2}{*}{Dataset} & \multicolumn{3}{c}{Conversations} & \multicolumn{3}{c}{Utterances} & \multirow{2}{*}{Classes} \\
\cmidrule(lr){2-4}\cmidrule(lr){5-7} 
& \multicolumn{1}{c}{Train}   & \multicolumn{1}{c}{Val}     & \multicolumn{1}{c}{Test} & \multicolumn{1}{c}{Train}  & \multicolumn{1}{c}{Val}    & \multicolumn{1}{c}{Test} & \\ 
\midrule
IEMOCAP & 100 & 20 & 31   & 4778 & 980 & 1622 & 6 \\
MELD    & 1038 & 114 & 280  & 9989 & 1109 &  2610 & 7 \\ 
EmoryNLP &713 & 99 &85 & 9934 & 1344 & 1328 & 7\\
\bottomrule
\end{tabular}
\caption{Statistics of the datasets.}
\label{tab:statics}
\end{table}
}

\begin{itemize}
\item \textbf{IEMOCAP} \cite{busso2008iemocap} consists of dyadic conversations between pairs of ten speakers, containing $153$ conversations and $7,433$ utterances. Each utterance is labeled as one of six emotion categories: happy, sad, neutral, angry, excited, and frustrated. 
% Following previous work (ref), we use the last dialogues in the training set for validation.
\item \textbf{MELD} \cite{poria2019meld} is a multiparty conversation dataset from Friends TV series, containing $1,433$ conversations, $13,708$ utterances. Each utterance is labeled as one of seven emotion categories: anger, disgust, fear, joy, neutral, sadness, and surprise.
\item \textbf{EmoryNLP} \cite{zahiri2018emotion} is also collected from Friends TV series, containing $97$ episodes, 897 scenes, and $12,606$ utterances. Each utterance is annotated as one of seven emotion types: neutral, joyful, peaceful, powerful, scared, mad, and sad.
\end{itemize}
% For evaluation, we follow prior works \cite{zhang2023dualgats, qin2023bert, fu2025laerc} and use the weighted average F1 score (w-F1) for the three datasets. %experiments. 

Following prior studies \cite{zhang2023dualgats, qin2023bert, fu2025laerc}, we use the weighted average F1 score (w-F1) to evaluate performance on all three datasets.
% for evaluation.
% most previous works
% overall accuracy (ACC) 看是否要加其他文献或者加其他的评估指标
% all datasets.
% weighted F1
% % \subsection{Evaluation Metrics}
% We follow most previous works and use weighted-average F1 score for evaluation.
% accuracy and
% 多下面的内容进行扩写 润色 优化，稍微增加文字数量，同时满足学术论文科学 严谨 合理的风格

% 对下面的内容进行纠错并优化，使得流畅，连贯，合理：
% 在llama-factory中lora_target: all
% lora_rank: 8
% 那么下面的句子应该怎么写

\subsection{Implementation Details}
% and adopt a similarity‑descending order, i.e., similar-first, for example ordering
We implement the proposed InitERC using LLaMA-Factory framework \cite{zheng2024llamafactory}. In this paper, we adopt LLaMA3.1-8B-Instruct \cite{grattafiori2024llama} as our LLM backbone. For in-context example selection, we set the number of examples $k$ to $5$. To enable parameter‑efficient fine‑tuning, we apply LoRA \cite{hulora} by inserting low-rank adapters in all linear layers, with the rank set to $8$. We employ AdamW optimizer \cite{KingmaB14}, with the learning rate of 1e-4, and the batch size of $8$. The maximum input length is set to $2048$ for IEMOCAP, and $1024$ for both MELD and EmoryNLP datasets. During inference, we use greedy decoding strategy to predict emotion labels. All experiments are conducted on a single NVIDIA RTX 4090 GPU.
% we apply LoRA \cite{hulora} by inserting low-rank adapters after self-attention layers, with the adapter dimension set to $8$. 
% dimension of adapters
% We employ AdamW optimizer to train InitERC for $3$ epochs. In details, we set the learning rate as $xx$, batch size as $4$.

% 对下面的baseline进行纠错 优化，有可能的话进行缩短，使得简洁，流畅，连贯，合理：
\subsection{Baseline Methods} %KET \cite{zhong2019knowledge}
To demonstrate the effectiveness of our InitERC framework, we compare InitERC with a range of conventional and LLM-based ERC baselines. For conventional ERC methods, we adopt both graph- and sequence-based methods. The graph-based methods include SKAIG \cite{li2021past}, DAG-ERC \cite{shen2021directed}, and DualGATs \cite{zhang2023dualgats}, which leverage psychological-knowledge-aware interaction graphs, directed acyclic graph structures, or dual graph attention networks to recognize emotion in conversation. The sequence-based methods include HiTrans \cite{li2020hitrans}, DialogXL \cite{shen2021dialogxl}, SPCL+CL \cite{song2022supervised}, and BERT-ERC \cite{qin2023bert}, which utilize hierarchical transformers, memory-augmented attention, contrastive learning, and pretrained language models to capture sequential and structural information within dialogues. For LLM-based methods, we compare with ChatGPT \cite{zhao2023chatgpt} using 3-shot prompting, InstructERC \cite{lei2023instructerc} with two-stage instruction tuning, BiosERC \cite{xue2024bioserc} which integrates speaker biographical knowledge to fine-tune LLMs, and LaERC-S \cite{fu2025laerc} that fine-tunes LLaMA2-7B using a two-stage learning to capture emotion dynamics.
\section{Results and Analysis}
\subsection{Main Results}
% To ensure a fair comparison with existing instruction‑tuned LLM methods, we also evaluate InitERC using LLaMA2‑7B as its backbone (denoted InitERC‑LLaMA2‑7B).
% consistently 
% From the table, we can observe that: 
Table~\ref{tab:main_results} reports the performance of our InitERC and other baseline methods on three datasets. Given that current LLM-based approaches utilize LLaMA2-7B as their backbone, we also adopt this LLM in InitERC for fair and consistent comparison. From the results, we have the following observations: (1) Methods based on instruction-tuned LLMs, such as BiosERC, InstructERC, and LaERC-S, outperform conventional graph-based models (e.g., DAG-ERC, DualGATs) and sequence-based models (e.g., DialogXL, HiTrans). This indicates the superior capacity of instruction tuning in modeling conversational dynamics compared to conventional architectures. (2) ChatGPT exhibits a notable performance gap compared to fine-tuned LLMs and even performs worse than most traditional methods. These results highlight that without fine-tuning, general-purpose LLMs are less capable of modeling the nuanced emotional dynamics required for ERC. (3) Our InitERC framework achieves new state-of-the-art performance, surpassing the strongest baseline (LaERC-S) by $20.25\%$, $9.44\%$, and $14.42\%$ on IEMOCAP, MELD, and EmoryNLP respectively. These significant improvements validate our one‑stage in‑context instruction‑tuning framework can capture the complex interplay between speaker characteristics, dialogue history, and emotional dynamics more effectively. (4) Even when adopting LLaMA2-7B as backbone, InitERC still demonstrates strong performance, outperforming all other baselines. This validates the rationality and superiority of our framework in harnessing LLMs potential for specialized ERC tasks.% than existing ERC methods
 % This confirms that the performance advantage of InitERC is not solely due to the model size, but stems from its carefully designed framework that leverages in-context examples and instruction tuning in a unified and robust manner.
% {
% \setlength{\tabcolsep}{1mm}  % 设置列之间的水平间距为 1mm（紧凑）
% \begin{table}[!t]
%     \centering
%     \renewcommand\arraystretch{1.25}
%     \begin{tabular}{lccc}
%         \toprule
%         Methods & IEMOCAP & MELD & EmoryNLP \\ 
%         \midrule
%         KET & 59.56 & 58.18 & 34.39 \\
%         HiTrans & 64.50 & 61.94 & 36.75 \\
%         DialogXL  & 65.94 & 62.41  & 34.73 \\
%         DAG-ERC &  68.03  & 63.65  & 39.02 \\
%         SPCL+CL & 69.74  & 67.25   & 40.94   \\
%         DualGATs & 67.68  & 66.90 & 40.69      \\ 
%         BERT-ERC & 71.70  & 67.11  & 39.84     \\ \hline
%         ChatGPT &  48.58   &   58.35   & 35.92 \\ 
%         InstructERC & 71.39  &  69.15    & 41.37   \\ 
%         BiosERC  &    69.02   &   68.72   &   41.44 \\ 
%         LaERC-S &  72.40   &  69.27     &   42.08 \\ \hline
%         InitERC-Llama2-7B & 80.50 & 76.66 & 55.11 \\
%         InitERC  &   \textbf{92.65}   &   \textbf{78.71} &  \textbf{56.50}    \\ 
%         \bottomrule
%     \end{tabular}
%     \caption{Experiment results on three datasets.}
%     \label{tab:main_results}
% \end{table}
% }
{
\setlength{\tabcolsep}{1mm}  % 设置列之间的水平间距为 1mm（紧凑）
\begin{table}[!t]
    \centering
    \renewcommand\arraystretch{1.25}
    \begin{tabular}{lccc}
        \toprule
        Methods & IEMOCAP & MELD & EmoryNLP \\ 
        \midrule
        % KET & 59.56 & 58.18 & 34.39 \\
        SKAIG & 66.96 & 65.18 & 38.88 \\
        DAG-ERC &  68.03  & 63.65  & 39.02 \\
        DualGATs & 67.68  & 66.90 & 40.69      \\ 
        HiTrans & 64.50 & 61.94 & 36.75 \\
        DialogXL  & 65.94 & 62.41  & 34.73 \\
        SPCL+CL & 69.74  & 67.25   & 40.94   \\
        BERT-ERC & 71.70  & 67.11  & 39.84     \\ \hline
        ChatGPT &  48.58   &   58.35   & 35.92 \\ 
        InstructERC & 71.39  &  69.15    & 41.37   \\ 
        BiosERC  &    69.02   &   68.72   &   41.44 \\ 
        LaERC-S &  72.40   &  69.27     &   42.08 \\ \hline
        InitERC-Llama2-7B & 80.50 & 76.66 & 55.11 \\
        InitERC  &   \textbf{92.65}   &   \textbf{78.71} &  \textbf{56.50}    \\ 
        \bottomrule
    \end{tabular}
    \caption{Experimental results on three datasets.}
    \label{tab:main_results}
\end{table}
}

\begin{figure*}[t]  % [t] 表示尽量放在页面顶部
    \centering
    \begin{subfigure}[b]{0.32\textwidth}  % 调整为 0.32 避免溢出
        \centering
        \includegraphics[width=\linewidth]{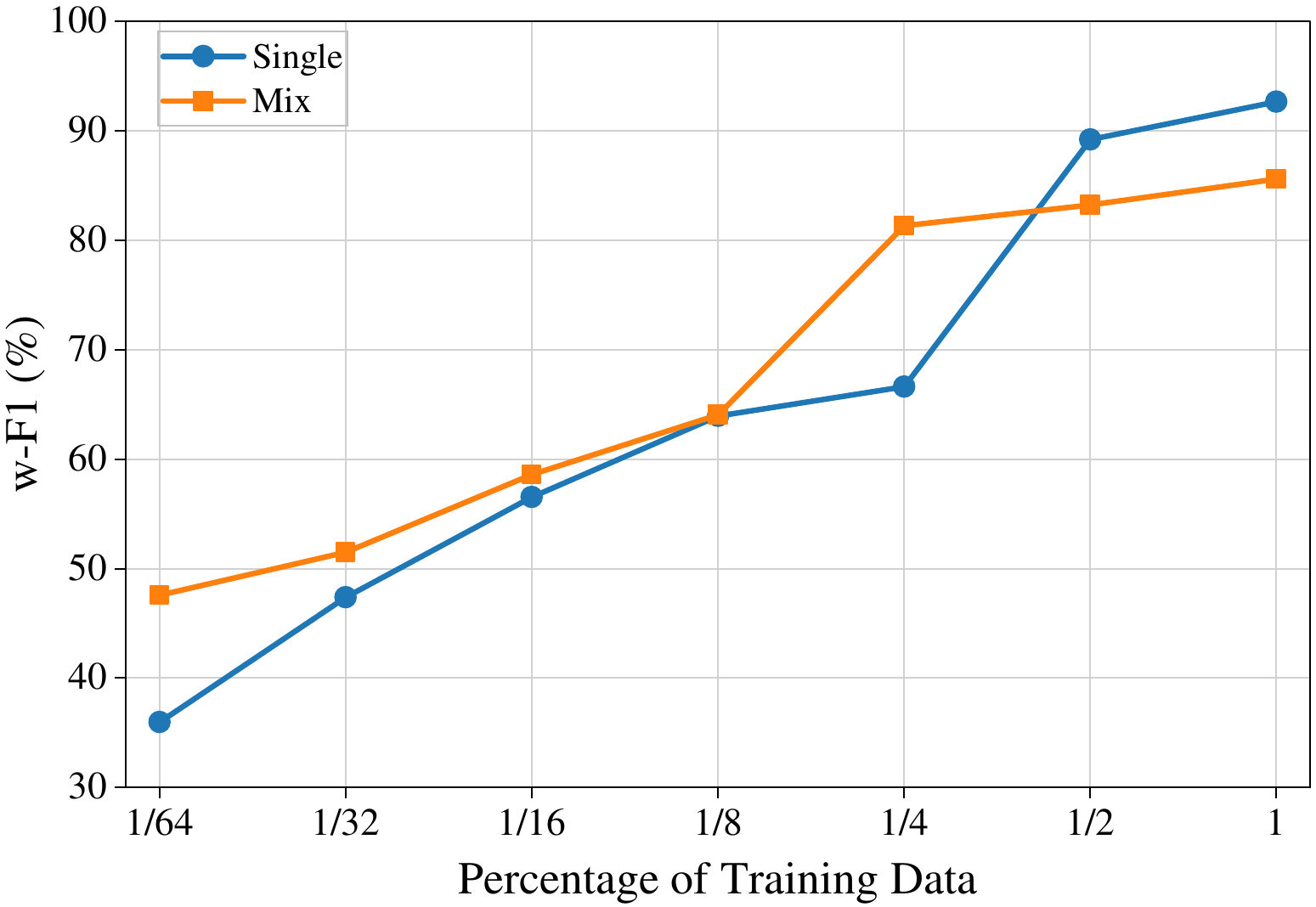}
        \caption{Experimental results on IEMOCAP.}
        \label{fig:iemocap}
    \end{subfigure}
    \hfill
    \begin{subfigure}[b]{0.32\textwidth}
        \centering
        \includegraphics[width=\linewidth]{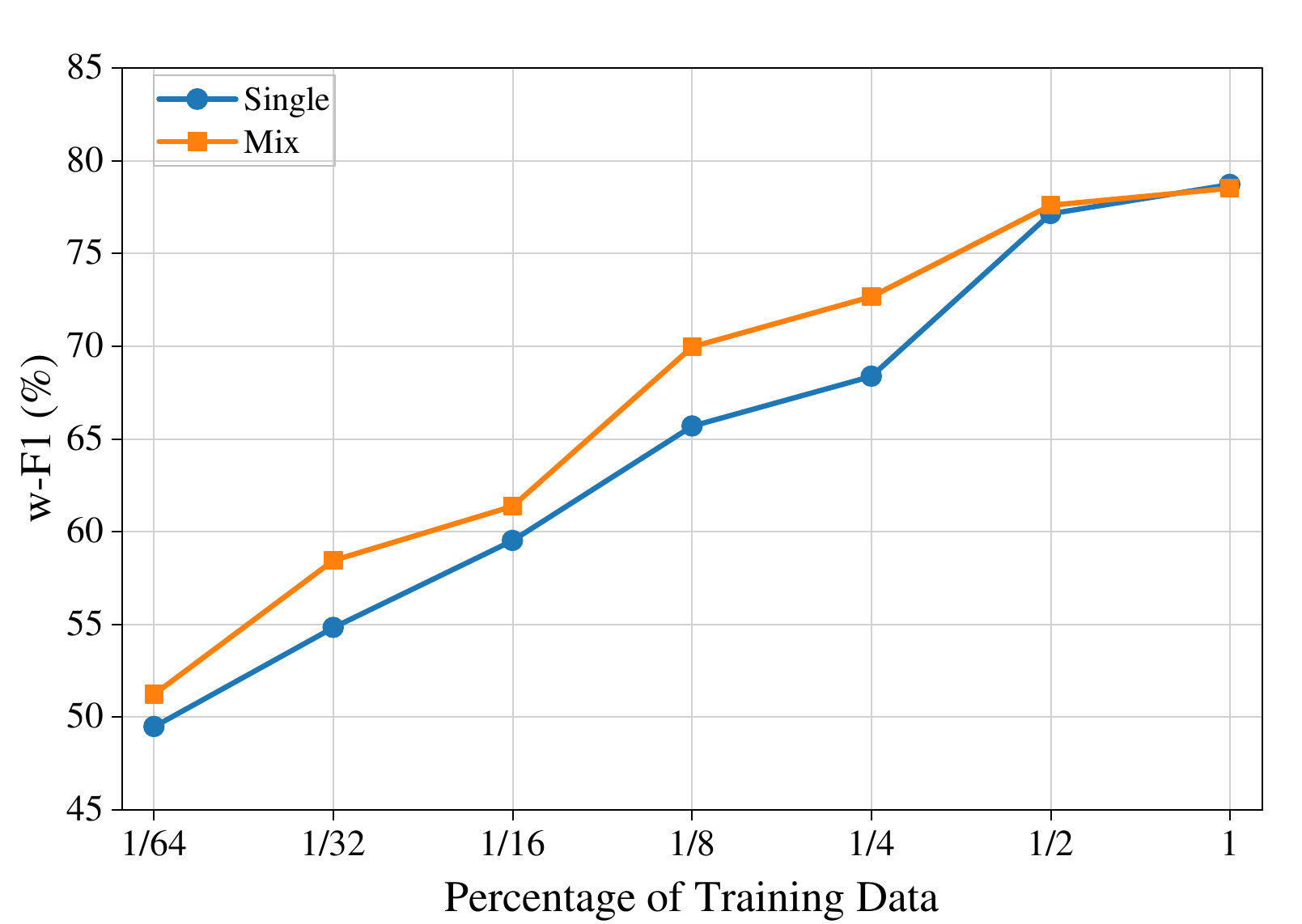}
        \caption{Experimental results on MELD.}
        \label{fig:meld}
    \end{subfigure}
    \hfill
    \begin{subfigure}[b]{0.32\textwidth}
        \centering
        \includegraphics[width=\linewidth]{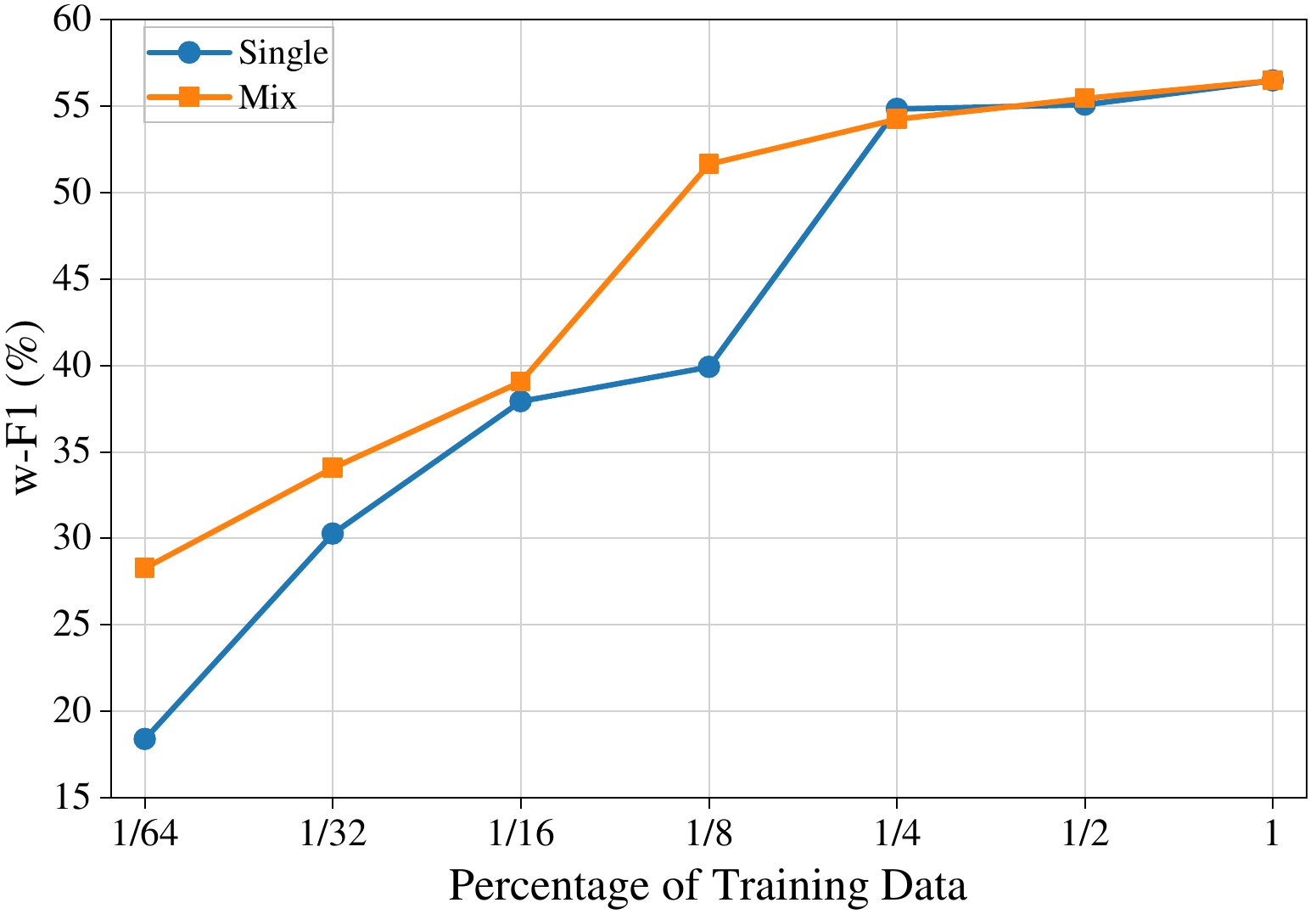}
        \caption{Experimental results on EmoryNLP.}
        \label{fig:emorynlp}
    \end{subfigure}
    \caption{Cross-dataset experimental results on three datasets. ``Single'' and ``Mix'' denote training on a single and mixed dataset, respectively. For both settings, we select data from each dataset at proportions of $1/64$, $1/32$, $1/16$, $1/8$, $1/4$, $1/2$, and $1$.}
    % in the ratios 
    % The ratios of $1/64$, $1/32$, $1/16$, $1/8$, $1/4$, $1/2$, and $1$ are selected data from each dataset.}
    \label{fig:mix-data}
\end{figure*} 
%``gpt-3.5-turbo'' 

\subsection{Ablation Study}

{
\setlength{\tabcolsep}{1mm}  % 设置列之间的水平间距为 1mm（紧凑）
\begin{table}[!t]
    \centering
    \renewcommand\arraystretch{1.25}
    \begin{tabular}{lccc}
        \toprule
        Methods & IEMOCAP & MELD & EmoryNLP \\ 
        \midrule
        InitERC  & \textbf{92.65} & \textbf{78.71}  & \textbf{56.50} \\
         \makecell[l]{w/o fine-tuning \\ \& in-context examples} & 36.50 & 20.87 & 25.38 \\
        w/o fine-tuning & 11.12 & 7.57 & 10.90 \\
        w/o in-context examples & 71.74 & 69.54  & 40.33 \\
         % \quad \quad 
        \bottomrule
    \end{tabular}
    \caption{Ablation results on three datasets.}%study 
    \label{tab:ablation_results}
\end{table}
}
We conduct an ablation study to investigate the impact of in-context examples and instruction tuning in our proposed InitERC framework. Table~\ref{tab:ablation_results} presents the ablation results. First, adopting a zero-shot prompting setting without both fine-tuning and in-context examples (i.e., w/o fine-tuning \& in-context examples) yields limited performance. This indicates that relying solely on the LLM's inherent zero-shot capabilities is insufficient to handle the complexity of ERC. Second, eliminating the fine-tuning stage while retaining the in-context examples (i.e., w/o fine-tuning) results in a severe performance collapse, even worse than the zero-shot prompting setting. This may be attributed to the fact that without fine-tuning, the presence of in-context learning might introduce inconsistencies or noise that the model is not able to interpret correctly, whereas a pure zero-shot prompt provides a more stable yet generic inference baseline. Third, removing in-context examples (i.e., w/o in-context examples) leads to a notable performance drop on all three datasets. This demonstrates that fine‑tuning alone cannot adequately learn the alignment among speaker characteristics, conversational context, and emotion states without demonstration examples. Overall, the results clearly demonstrate that InitERC—which jointly incorporates in-context examples and instruction tuning—significantly outperforms all ablation variants, which validates the design rationality of our framework.

\subsection{Cross-Dataset Robustness Analysis}
To validate our InitERC's robustness capability, we perform cross-dataset evaluation experiments. First, we sample equal proportions from the training sets of IEMOCAP, MELD, and EmoryNLP, and merge them into a mixed dataset. We then fine-tune InitERC on the mixed dataset and inference separately on the test sets of the three original datasets. For comparison, we also report the results of InitERC when both training and inference on the original datasets. Following previous work \cite{lei2023instructerc,fu2025laerc}, we map the emotion labels of the three datasets into a unified label space before mixing to ensure label consistency across datasets. 

% 补充完整 分三点描述实验结果，最后对鲁棒性进行总结
Figure~\ref{fig:mix-data} presents cross-dataset experimental results on three datasets. We can observe that InitERC generally benefits from training on the mixed dataset under low-resource conditions ($\leq 1/8$), demonstrating robust generalization when data is scarce due to the supply of complementary conversational cues. Moreover, as the training size increases, the gap between InitERC trained on mixed- and single-dataset narrows on MELD and EmoryNLP, yet the mix-trained InitERC remains competitive, indicating its robustness across varying data scales and heterogeneous distributions. In contrast, mixed training underperforms single‐dataset training on IEMOCAP at $1/2$ and full data scales, likely stems from the mismatch between IEMOCAP's dyadic, acted dialogues and the multi-party interactions in MELD and EmoryNLP. Overall, InitERC shows strong cross-dataset robustness, particularly in low-resource scenarios and on structurally compatible datasets. However, when the target domain has distinctive discourse structures, indiscriminate mixing can undermine robustness once sufficient in-domain data is available.

\subsection{Generalization Analysis across LLMs}
% To evaluate InitERC's generalization capabilities across different LLMs
To evaluate the generalization capability of InitERC across different LLMs, we conduct experiments with another open source LLM, namely Qwen2.5-7B-Instruct \cite{yang2025qwen3}. Furthermore, we instruction-tune each LLM using LoRA without any in-context examples, serving as comparison baselines. Table~\ref{tab:cross-llms} presents the results, showing that InitERC consistently outperforms the LoRA-only baselines across both LLM backbones. Specifically, InitERC-LLaMA achieves the highest performance on all datasets, significantly surpassing its LoRA-LLaMA baseline. Similarly, InitERC-Qwen demonstrates substantial improvements over LoRA-Qwen across all datasets. It is worth noting that InitERC with LlaMA performs slightly better than with Qwen, primarily due to LLaMA-3.1-8B-Instruct's larger model capacity and broader training data, which provide richer priors for emotion recognition. These findings highlight that our in-context instruction tuning framework is effective and generalizable across different LLMs, enabling stronger alignment of speaker identity, contextual cues, and emotion states.

{
\setlength{\tabcolsep}{1mm}  % 设置列之间的水平间距为 1mm（紧凑）
\begin{table}[!t]
    \centering
    \renewcommand\arraystretch{1.25}
    \begin{tabular}{lccc}
        \toprule
        Methods & IEMOCAP & MELD & EmoryNLP \\ 
        \midrule
        LoRA-Qwen  &   71.28   &   68.95 &  38.74   \\ 
        % LoRA-Mistral  &   -   &   - &  -    \\ 
        LoRA-LLaMA  &  71.74   &  69.54 &  40.33    \\ \hline
        InitERC-Qwen  &   91.05   &  77.52  &  56.35   \\ 
        % InitERC-Mistral  &   -   &   - &  -    \\ 
        InitERC-LLaMA  &   92.65   &   78.71 &  56.50    \\ 
        \bottomrule
    \end{tabular}
    \caption{Experimental results across different LLMs.}
    \label{tab:cross-llms}
\end{table}
}

% 对下面的内容进行英文翻译 并进行优化 修改 润色。总结为一段话，需要科学 严谨 合理 流畅，满足学术论文的要求。

\subsection{Further Analysis}
To further investigate the impact of in-context examples, we conduct comprehensive study of three key factors in in-context example selection including retrieval strategy, example ordering, and the number of examples. 

\subsubsection{Impact of Retrieval Strategy}
To investigate the impact of retrieval strategy, we conduct experiments on different retrievers: Random, BM25 \cite{robertson2009probabilistic}, SBERT \cite{reimers2019sentence}, and Contriever \cite{izacardunsupervised}. As shown in Table~\ref{tab:retriever}, the performance of InitERC varies considerably across different retrieval strategies. This highlights  the critical role of retrieval strategy in our InitERC framework, where effective retrieval of in-context examples can substantially  improve emotion recognition. Therefore, careful design retrieval methods warrants further attention and in-depth investigation. Moreover, future work could explore the potential of jointly training the retriever and the emotion recognition model to further  harmonize example selection with emotion inference.
% 因此设计值得关注，需要进一步研究，另外在未来可以考虑共同训练xx，以xx。
% with different retrieval strategies有较大的差距. 
% The result demonstrates the critical role of retrieval strategy in our InitERC framework, where effective retrieval of in-context examples can substantially enhance emotion recognition performance by providing xx. 因此采用哪种检索策略值得关注，需要进一步研究，另外在未来可以考虑共同训练xx，以xx。
% demonstrating 
% The result demonstrates the critical role of retrieval strategy in our InitERC framework, where effective retrieval of in-context examples can substantially enhance emotion recognition performance by providing xx. 
% retrieval strategy has a substantial influence on the performance of InitERC across all three datasets. Among them, 
% These results demonstrate the critical role of retrieval strategy in the InitERC framework, where effective retrieval of in-context examples
% —particularly through dense retrievers like Contriever—can substantially enhance emotion recognition performance by providing more relevant and informative demonstrations.
% the performance of InitERC with different retrieval strategies有较大的差距，因此采用哪种检索策略值得关注，需要进一步研究，另外在未来可以考虑共同训练xx，以xx。
% 检索策略值得关注 , including
{
\begin{table}[!t]
    \centering
    \renewcommand\arraystretch{1.25}
    \begin{tabular}{lccc}
        \toprule
        Retriever & IEMOCAP & MELD & EmoryNLP \\ 
        \midrule
         Random & 92.25 & 69.24 & 52.09 \\
         BM25 & 90.52 & 76.41 & 56.22 \\
         SBERT & 88.74 & 75.27 & 53.72 \\
         Contriever & 92.65 & 78.71 & 56.50 \\
        \bottomrule
    \end{tabular}
    \caption{Impact of retrieval strategy on InitERC performance.}%study 
    \label{tab:retriever}
\end{table}
}
%``gpt-3.5-turbo'' 
\subsubsection{Impact of In-Context Example Ordering}
\begin{figure}[t]
\centering
\includegraphics[width=1.0\columnwidth]{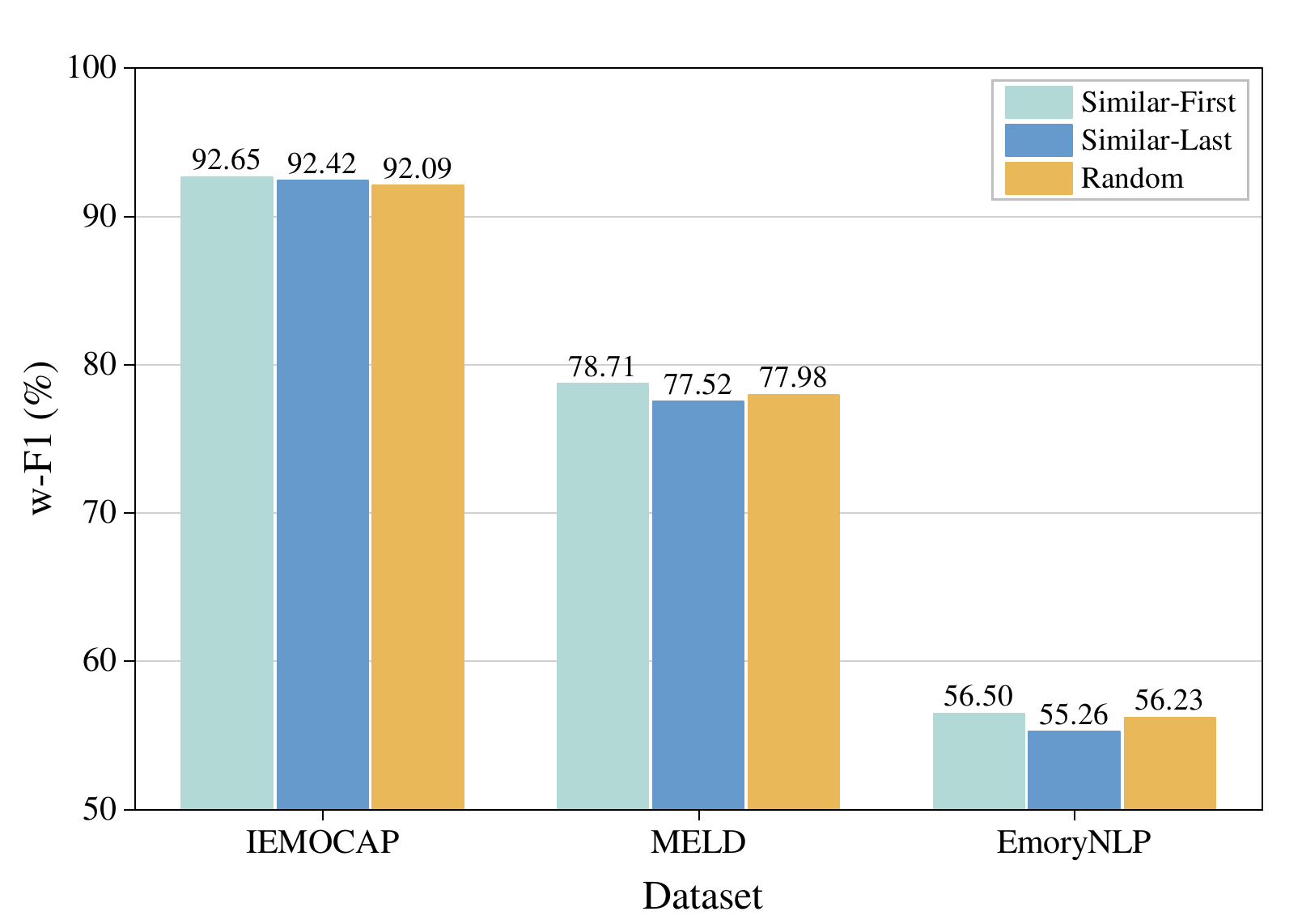} % Reduce the figure size so that it is slightly narrower than the column. Don't use precise values for figure width.This setup will avoid overfull boxes.%width=0.85\textwidth
\caption{Impact of in-context example ordering on InitERC performance.}
\label{ordering}
\end{figure}

To investigate how the in-context example ordering impacts the efficacy of InitERC, we carry out experiments on various orderings: ``Similar-First'', where the most similar examples are positioned at the beginning; ``Similar-Last'', where the most similar examples are placed at the end; and ``Random'', where the order of examples is randomized. As illustrated in Figure~\ref{ordering}, the performance differences across ordering strategies are relatively marginal on all datasets, indicating that the ordering of in‑context examples has only a minimal influence on the performance. The finding demonstrates that InitERC is inherently robust to the sequence of demonstration examples, and specifically designed ordering schemes may be unnecessary, thereby simplifying the construction of in-context prompts.

% 模仿上述两个分析，并根据上述图表，对下面的内容补充完整，并进行优化 润色，实现对结果的分析。总结为一段话，需要科学 严谨 合理 连贯 自然，满足学术论文的要求。
% the number of in-context examples
\subsubsection{Impact of Number of In-Context Examples}
\begin{figure}[t]
\centering
\includegraphics[width=1.0\columnwidth]{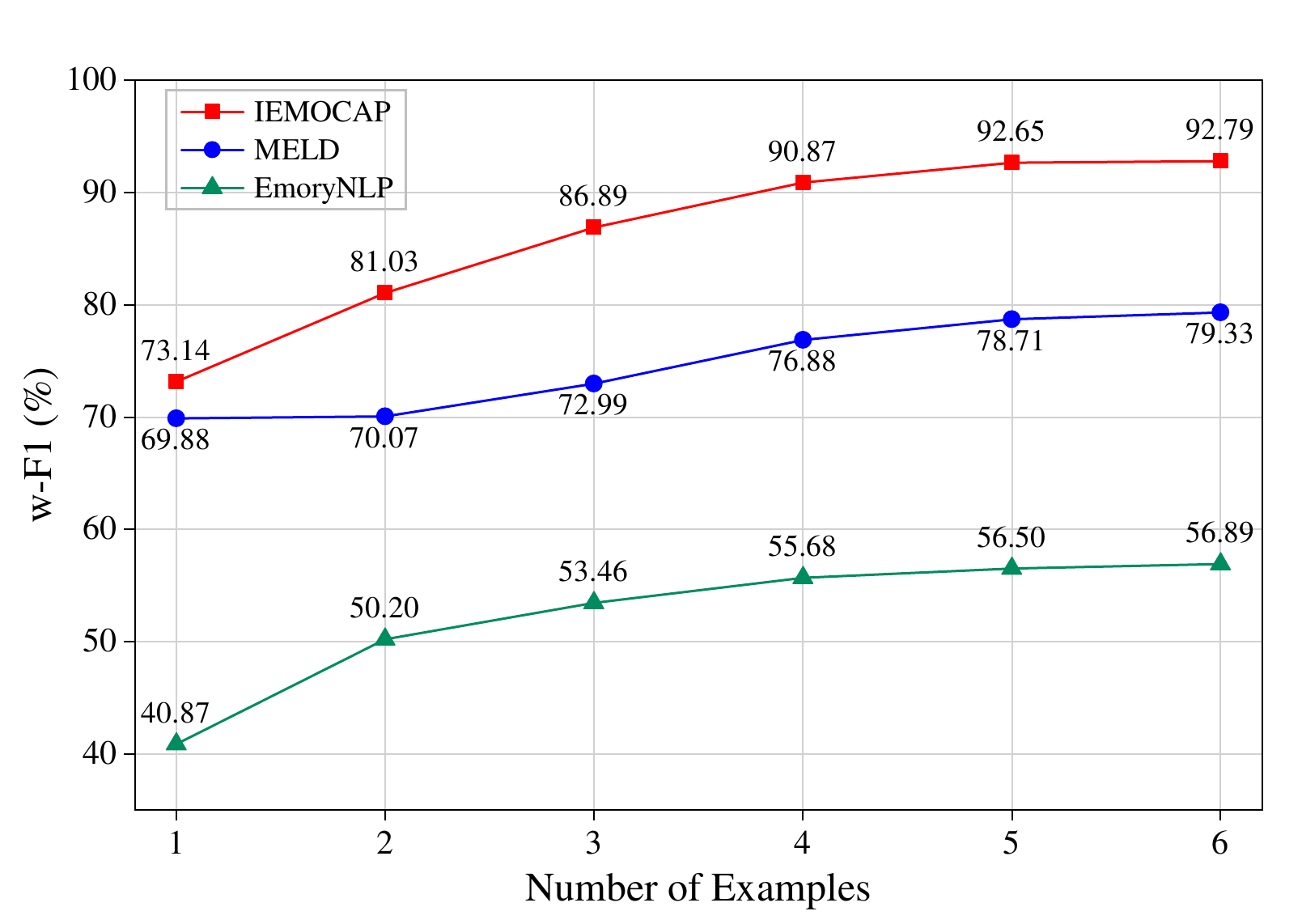} % Reduce the figure size so that it is slightly narrower than the column. Don't use precise values for figure width.This setup will avoid overfull boxes.%width=0.85\textwidth
\caption{Impact of number of in-context examples on InitERC performance.}
\label{number}
\end{figure}

To investigate the impact of the number of in-context examples $k$ on the performance, we vary $k$ from $1$ to $6$. From Figure~\ref{number}, we can observe that across all three datasets, as $k$ increases, the performance of InitERC generally shows an upward trend, although the rate of improvement gradually saturates. This result demonstrates that increasing the number of in-context examples contributes positively to our InitERC performance, particularly in the early phase, but with diminishing returns beyond a certain point. This finding highlights the practical trade-off between computational efficiency and performance gains when scaling in-context example size, and further implies that a moderate number of well-retrieved examples is sufficient for effective instruction tuning in ERC. 
% may suffice for effective instruction tuning.
% As shown in Figure~\ref{number}, 
% From Figure~\ref{number}, we can observe xx
% Figure~\ref{number} shows the performance of our InitERC with different  number of in-context examples.
% From Figure~\ref{number}, we can observe xx
% As depicted in Figure xx, 

% \subsection{Robustness Analysis}

% \subsection{Robustness of InitERC}

% 纠正错误 优化，使得科学 严谨 合理 流畅 自然，满足学术论文的要求
\section{Conclusion}
In this paper, we propose InitERC, a one‑stage instruction tuning framework for ERC that enables LLMs to learn the alignment among speaker identity, conversational context, and emotional state from in-context examples. InitERC contains demonstration pool construction, in-context example selection, prompt template design, and in-context instruction tuning: it first creates a candidate set of examples, retrieves the most relevant demonstrations, integrates the task description, target input, and selected examples into a unified prompt format, and finally fine-tunes LLMs using the constructed instruction prompts. Experimental results on three benchmark datasets show that the proposed framework outperforms both conventional and LLM-based ERC baselines, demonstrating the effectiveness of one-stage in-context instruction tuning. Moreover, our comprehensive analysis of retrieval strategy, example ordering, and the number of in-context examples provides useful suggestions for optimizing example selection in InitERC. We hope these findings can offer some insights for future research on in-context instruction tuning for ERC.

\bibliography{aaai2026}


% \input{checklist}

\end{document}